# Fuzzy-Depth Objects Grasping Based on FSG Algorithm and a Soft Robotic Hand

Hanwen Cao, *Student Member*, *IEEE*, Junda Huang, Yichuan Li, Jianshu Zhou*, *Member*, *IEEE*, and Yunhui Liu, *Fellow*, *IEEE*

*Abstract*—Autonomous grasping is an important factor for robots physically interacting with the environment and executing versatile tasks. However, a universally applicable, cost-effective, and rapidly deployable autonomous grasping approach is still limited by those target objects with fuzzy-depth information. Examples are transparent, specular, flat, and small objects whose depth is difficult to be accurately sensed. In this work, we present a solution to those fuzzy-depth objects. The framework of our approach includes two major components: one is a soft robotic hand and the other one is a Fuzzy-depth Soft Grasping (FSG) algorithm. The soft hand is replaceable for most existing soft hands/grippers with body compliance. FSG algorithm exploits both RGB and depth images to predict grasps while not trying to reconstruct the whole scene. Two grasping primitives are designed to further increase robustness. The proposed method outperforms reference baselines in unseen fuzzy-depth objects grasping experiments (84% success rate).

*Index Terms*- Robotic grasping, soft robotic hand, fuzzy-depth objects, specular objects.

## I. INTRODUCTION

As robots going out of laboratories and factories and coming into human daily life, robotic grasping is prominent for its physical interaction towards human beings and their environment. Efficient, reliable, and safe grasping capability of robots provides the potential for their broad applications [1-3].

In an autonomous grasping task, successfully sensing the geometry of grasping targets are prerequisites of planning grasps. For this reason, commercially available RGBD cameras are popular for depth information acquisition thanks to their cost-effectiveness, informative sensing, and convenient deployment [13, 21]. For most off-the-shelf RGBD cameras, specific patterns are projected from projectors and captured by imagers [14] to acquire depth information. However, the depth acquisition is highly influenced by properties of target surfaces and lighting conditions. For example, the depth of transparent, reflective, and small/flat objects (i.e., objects having little heights) are difficult to be accurately sensed by RGBD cameras.

There is a nuance in difficulties of depth acquisition between different objects. For transparent and specular objects, the optical pattern reflected from the surface is either too weak, too strong, or deviated from the original path, resulting wrong or missing depth estimation [4, 6]. For small and flat objects, the difficulty lies in the minor depth difference between the surface and background compared with sensor resolution. Such a minor depth difference raises requirements on robot control accuracy [23] or extra sensors [22].

The above-discussed two difficulties can be generally classified as a fuzzy-depth problem. Such fuzzy-depth objects are popular in grocery stores, logistics, and many other scenarios suitable for robotic system deployment. Therefore, the widespread application of depth-based grasping is still limited, and it is valuable to address this challenge by a universally applicable, cost-effective, and rapidly deployable method.

There are three ways to tackle fuzzy-depth object grasping. One is from sensing perspective, improving or modifying the current depth sensing approaches. Efforts are being reported recently that [9] demonstrates grasping transparent objects by scene reconstructing, while [10] uses a light field camera to better estimation poses of transparent objects. At the same time, [12] detects grasp poses in transparent clutter by plenoptic imaging. However, the development and refinement of those sensors require a long iteration cycle, and currently the cost of which is relatively large for daily application.

The second way is by integrating extra sensors, such as tactile or proximity sensors, making up for the shortcomings of depth sensing on fuzzy-depth objects. [8] improves state-of-the-art object grasping methods on transparent and specular objects by transfer learning with the help of gripper contact detection. [11] uses a non-contact pre-touch sensor to sense and explore. However, these strategies solve the issue at the cost of system complexity.

Soft robotic grasping developed in recent years attracts huge attention for their reliability, adaptability, and robustness enabled by inherent compliance, cost-effectiveness, and easy-to-control features [5, 7]. In this work, we proposed a grasping solution for fuzzy-depth objects based on soft robotic hands. The framework of our grasping approach comprises of two major components: one is a soft robotic hand, and the other is fuzzy-depth soft grasping (FSG)

The work was supported in part by Hong Kong RGC via the TRS project T42-409/18-R and 14202918, the Hong Kong Centre for Logistics Robotics, CUHK T Stone Robotics Institute, and in part by the CUHK Hong Kong-Shenzhen Innovation and Technology Research Institute (Futian).

Corresponding author: Jianshu Zhou (jianshuzhou@cuhk.edu.hk)
HW Cao, JD Huang, YC Li, JS Zhou, and YH Liu are with the T Stone Robotics Institute, the Department of Mechanical and Automation Engineering, The Chinese University of Hong Kong, HKSAR, China (email: {hwcao,jdhuang99}@link.cuhk.edu.hk) .
JS Zhou and YH Liu are also with the Hong Kong Center for Logistics Robotics (HKCLR) .

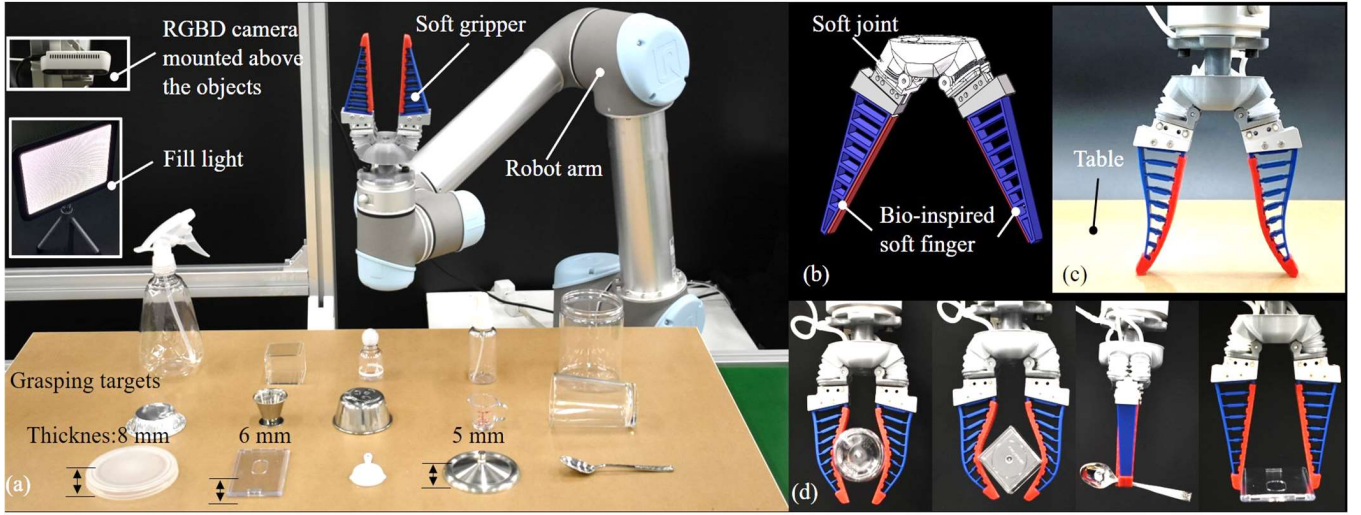

Figure 1. (a) The overall grasping system setup. A group of household items are selected as grasping targets, including transparent, specular, flat and small objects.. (b) Tstone-Soft (TSS) gripper construction. (c) TSS gripper passive deformation towards the environment. (d) TSS gripper grasping adaptability towards objects in various shapes including circular, quadrangular, and flat targets.

algorithm. The former can be replaced by most existing soft hands/grippers with body compliance. The FSG algorithm takes a pair of RGBD images as input and feeds it into a fully convolutional network (FCN). 3D position and planar orientation of the best grasp are predicted. We proposed two grasping primitives to further increase the robustness of grasping universal objects including transparent/reflective and small/flat objects. Experimental results show that our proposed method is able to grasp fuzzy-depth objects robustly using a soft robotic hand.

## II. Grasping System Overview

### A. Grasping System Overview

The overall system setup is illustrated in Fig 1(a) where a soft gripper is mounted at the tip of a collaborative robotic arm (UR5). A commercially available RGBD camera (RealSense D435i) is fixed about 1 m above the workspace. Two kinds of common fuzzy-depth objects are prepared. One is transparent and specular objects made in glass, stainless steel, and plastic; the other is small and flat objects less than 10 mm in height. We also note that some items have multiple attributes, such as being transparent and specular and flat. An extra fill light is used to adjust lighting condition as shown in Fig 1(a).

### B. Soft Robotic Hand and Its Characteristics

We designed a typical soft gripper, TStone-Soft (TSS) gripper, to execute compliant grasping. The construction of the gripper is illustrated in Fig 1(b), where the gripper is composed of two soft joints and two bio-inspired soft fingers controlled by a pneumatic actuator [15, 16].

There are two specific compliance/adaptability of soft gripper which benefits fuzzy-depth objects grasping. One is the passive deformation against the environment as illustrated in Fig 1(c) [17, 18], which enables the algorithm to exploit environmental constraints when grasping flat objects; the other is the adaptability against grasped objects as shown in Fig 1(d) [19]. TSS gripper can be replaced by most soft robotic hands with these two characteristics [20].

## III. Fuzzy-Depth Soft Grasping (FSG) Algorithm

In this section, we introduce our fuzzy-depth soft grasping (FSG) algorithm. The algorithm is designed for grasping fuzzy-depth items. Using RGB and inaccurate depth information captured by the camera, the algorithm outputs a specified grasping primitive to perform a robust grasp.

### A. Problem Statement

Our goal is to detect a grasp
$$\boldsymbol{g}_w = (x_w, y_w, z_w, \theta_w, w_w, q), \quad (1)$$
where $(x_w, y_w, z_w, \theta_w)$ is the gripper's center Cartesian position and rotation around $z$ axis. $w_w$ is the gripper's width. A scalar quality measure $q$ represents the chance of grasp success. The grasp is executed perpendicular to the $x-y$ plane. We assume access to RGB and depth image pair, $(\boldsymbol{I}_c, \boldsymbol{I}_d)$, where isolated objects are put on a planar surface. We want to detect grasps in the images. A grasp in the image is described by
$$\boldsymbol{g}_p = (x_p, y_p, \theta_p, w_p, q), \quad (2)$$
where $(x_p, y_p)$, $\theta_p$ and $w_p$ are grasp's center, planar rotation and gripper width in image coordinate. It is assumed that intrinsic and extrinsic parameters of the camera are known. Then $\boldsymbol{g}_p$ can be converted to $\boldsymbol{g}_w$ by
$$\boldsymbol{g}_w = \boldsymbol{T}_p^w(\boldsymbol{g}_p, z), \quad (3)$$
where $\boldsymbol{T}_p^w$ is the transformation from image coordinates to the world frame, and $z = \boldsymbol{I}_d(x_p, y_p)$ is the measured depth at the grasp point.

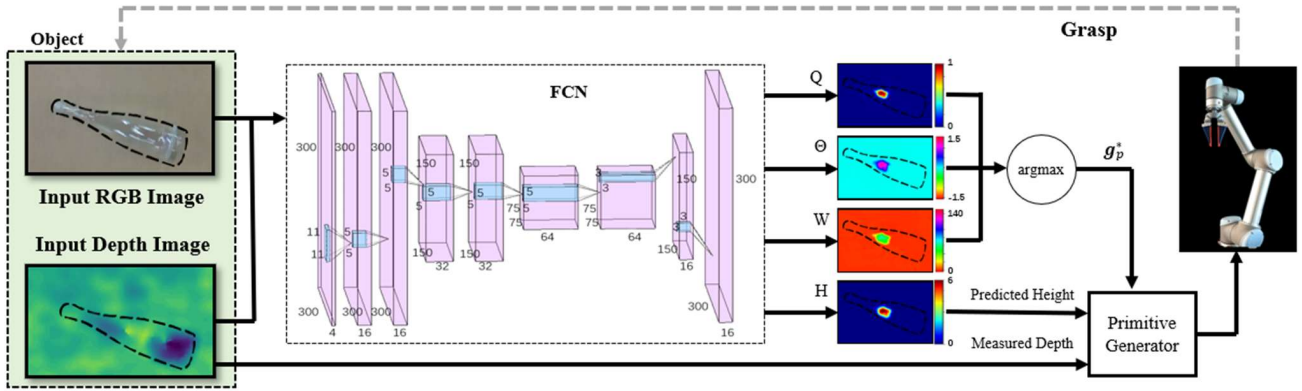

Figure 2. The overview of grasping pipeline. A fully convolutional network takes inpainted depth and color images and outputs 4 maps: the grasp quality map $Q$, grasp angle map $\Theta$, grasp width map $W$, and height map $H$. Each pixel in the first three maps represent a grasp pose $g_p$. The last map predicts object height at each grasp point. By jointly considering predicted object height and values in depth images, a robust grasp is generated.

Depth images are essential input for grasp synthesis methods [24-27]. Although these depth-based grasp detection methods have shown great success on standard grasping benchmarks such as Cornell dataset [28], Jacquard dataset [29], and [30] by harnessing the power of convolutional neural networks (CNNs), they are not guaranteed to work for every real-world scenario. We observe that though most objects can be sensed precisely by commercial RGBD cameras, some objects (e.g., transparent objects, reflective objects, flat objects, etc.) result in noisy or missing data in the corresponding depth images. Specifically, missing depth data are often caused by specular reflections on objects' surface; inaccurate depth estimation is caused by the distorted view of the background surfaces behind a transparent object; flat objects partially or completely blend into the background surface in the depth images due to the sensor noise.

In a word, due to object materials or sensing noise, the depth images could be highly uncertain and unpredictable, which makes it challenging to plan the grasps.

On the one hand, inference grasps would become difficult because depth images do not reflect the real geometry of objects to be picked. These objects sensed with uncertain depth measurement confuse depth-based grasping methods because the corresponding depth images provide fewer (usually for transparent and flat objects) or misleading (usually for specular objects) features for grasp synthesis.

On the other hand, even though a grasp can be detected in 2D images, the robot has vague clues of where to execute the grasp in the world coordinate because of the missing or incorrect data in the 3D measurement. For ordinary objects, after detecting a grasp in the depth image, the Cartesian position of the gripper's center when grasping can be obtained by Eq (3). However, for special objects described above, $z$ itself has huge uncertainty. Therefore, we need to find a solution to this challenge.

### B. FSG Algorithm Overview

Different input modalities (e.g., depth vs RGB) have complementary advantages [8]. We observe that many transparent, specular, and flat objects can be viewed clearly in RGB images rather than depth images, while ordinary objects provide more information for RGBD images. Besides, we observe in experiments that using RGB images improves the performance of detecting high-quality grasps when depth measurement is specious. Based on the above observations, we decide to make use of the ability of neural networks to detect grasps and avoid the confusion caused by uncertain depth data by inputting a pair of RGB and depth images to the network.

Depth measurement is crucial for not only synthesis grasps but also transfer grasps to desired poses in the world coordinate when executing them. To determine the depth $z$ of the gripper when grasping, instead of directly reading from unreliable depth images, we wish to get more insight into the object to be grasped from RGB images. We assume that the height of an regular geometry object is a function of its appearance as

$$f: (I_c, I_d) \to h, \qquad (4)$$

where $h$ is the relative height of objects at the grasp point to the supporting planar surface. We show in experiments that the deep neural network is able to approximate $f$ by simply glancing at a pair of RGB and uncertain depth images.

By predicting $h$, we could fill the information gap when depth measurement is uncertain, which benefits robust grasping with further help of soft grasping primitives in Sec. III(C).

To this end, we design a grasping pipeline and a deep neural network to output not only the quality and pose of a grasp but also the relative height of the object at the center of the grasp to the plane surface beneath.

### C. Grasping Pipeline and Soft Grasping Primitives

The grasping pipeline consists of three stages: image pre-processing, network prediction, and soft grasping primitives, as shown in Fig 2.

**Pre-processing:** We acquire a pair of RGB and aligned depth images. The depth image is applied to spatial edge-preserving filters [31], and temporal filters [32]. The RGB image and depth image is then cropped to 300 300 pixels to suit the input of our neural network. We further inpaint invalid depth values using OpenCV [33].

**Pixel-wise prediction:** Our network output multiple maps which jointly determine a grasp. Similar to GG-CNN [39], our network produces four $300 \times 300$ maps that represent the quality $Q$, planar angle $\Theta$ (one map for $\sin 2\theta$ and one map for $\cos 2\theta$), gripper width $W$ of grasps executed at each pixel. The best grasp can be obtained from

$$(x_p^*, y_p^*) = argmax(Q) \quad (5.a)$$
$$w_p^* = W(x_p^*, y_p^*) \quad (5.b)$$
$$\theta_p^* = \Theta(x_p^*, y_p^*) \quad (5.c)$$

To predict the height of the object at each grasp point $(x_p, y_p)$, our network also outputs another map $H$ representing the relative height. Then $h$ is estimated by

$$h^* = H(x_p^*, y_p^*) \quad (6)$$

As shown by [34], smoothing the output quality map can lead to superior grasps compared with the original quality map. Thus, before looking for the maximum in the quality map, we filtered $Q$ by a Gaussian kernel to smooth the local maxima. In case of multiple maxima exist and sit close to each other, we perform Connected Component Labeling [35] on all maxima and get the geometry center of each connected component.

**Soft grasping primitives:** Due to the uncertain depth measurement, we do not read the gripper's grasping depth $z^*$ from $I_d$ at the best grasp point. Instead, we design a set of primitives to handle the uncertainty in depth images. According to the range of predicted height of each grasp, the soft gripper behaves by jointly consider the predicted height from network output and measured depth in depth images as presented in the following.

**Grasping primitive 1:** Grasping Flat/Small Objects. We first calculate the measured height of the grasp point by

$$h_m^* = I_d(x_p^*, y_p^*) - d_t, \quad (7)$$

and get the predicted height from Eq (6), assuming the depth of the plane $d_t$ can be obtained. If both the measured height and the predicted height at the best grasp point are smaller than a height $h_c$, i.e., $max(h^*, h_m^*) < h_c$, first primitive is selected. Specifically, the gripper first moves vertically towards the supporting plane to $(x_w^*, y_w^*, z_1, \theta_w^*, w_w^*)$, where $z_1$ is a constant height, such that the gripper's fingers have tight contact with the supporting surface, and then closes its fingers while moving vertically up.

**Grasping primitive 2:** Grasping Normal-sized Objects. For normal height prediction $max(h^*, h_m^*) \geq h_c$, the gripper adjusts grasping height according to objects' height. Meaningly, the gripper first moved vertically towards the plane and then close its fingers at $(x_w^*, y_w^*, z_2, \theta_w^*, w_w^*)$, where $z_2 = min\{z_1 + h^*, I_d(x_p^*, y_p^*)\}$. This primitive is helpful to prevent hard collisions between the robot and objects.

We note that our method only needs a rough range of objects' height at the grasp points, rather than precisely 3D reconstruction of the whole objects, to determine the appropriate behaviors of the soft gripper. Besides, since our soft gripper is compliant, it can achieve robust grasping given a wide range of grasping height, which reflects the synergy between the grasping primitives and the soft gripper.

### D. Dataset

Due to the lack of existing benchmark datasets for special objects that cause uncertain depth sensing[8], we collect our own training set.

Our training set contains 90 pairs of data over 10 household items. 9 out of 10 items are either transparent or specular or flat, and 1 item is opaque. All the RGB and depth images are captured by an Intel RealSense D435i camera. The depth images have been passed through spatial [31] and temporal filters [32] and OpenCV [33] inpaint offline.

Several successful grasps are annotated during grasp trials as labels, while failure grasps are neglected. We use the commonly used rectangle representation for each grasp [28]. The relative height of the object at successful grasp points is measured manually.

In order to train our neural network to output pixel-wise grasp and height prediction, we adopt the same labeling method as [24]. The rectangle representations of all grasps are converted to image masks which correspond to the position of the center of the gripper. This image mask is further used to update our training data. Height labels are treated in the same manner.

Input images and labels are normalized during training. In addition, we augment the dataset with random crops, zooms, and rotations online to create more training examples.

### E. Network Architecture

Fully convolutional networks have shown great advantages in computer vision tasks, such as semantic segmentation [36] and robotics grasping. Our neural network is designed based on the fully convolutional grasp detection network GG-CNN2 [24], which further uses dilated convolutional layers [36] to provide improved performance. To achieve robust performance in challenging perception situations, we keep the basic structure but expand filter sizes for the 3rd and 4th convolutional layers from 16 to 32 and for the dilated convolutional layers from 32 to 64, as shown in Fig 2. Though this may slightly increase the complexity of the network, we will show in experiments that the inference time of a pair of input images is satisfying.

To further improve the performance of our neural network, we leverage the advantage of over-parameterization [37]. By augmenting a convolutional layer with a depth-wise convolution that convolves each input channel separately, CNNs achieve faster training and boosted performance but maintain forward inference time [37].

The inputs of our neural network are 4-channel $300 \times 300$ RGBD images, and the outputs of our network are corresponding maps $Q$, $\Theta$, $W$, and $H$. The Mean Squared Error (MSE) loss is performed pixel-wise to learn the mapping from inputs to outputs.

We train our network on 90% of our training dataset, and keep 10% for evaluation. Training is optimized by Adam with a batch size of 32.

## IV. Experiments

To evaluate our proposed grasping approach, several experiments were carried out. Firstly, we performed grasping experiments on isolated fuzzy-depth objects. Secondly, we

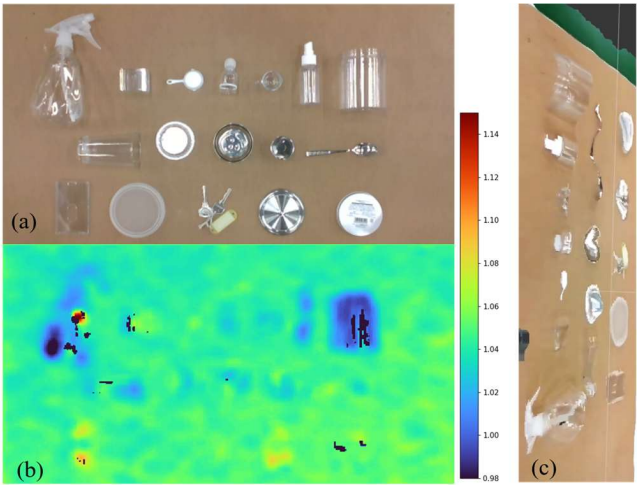

Figure 3. RGBD camera output of prepared fuzzy-depth objects. (a)(b) are respectively RGB image and depth image from top view. (c) Point cloud from inclined view.

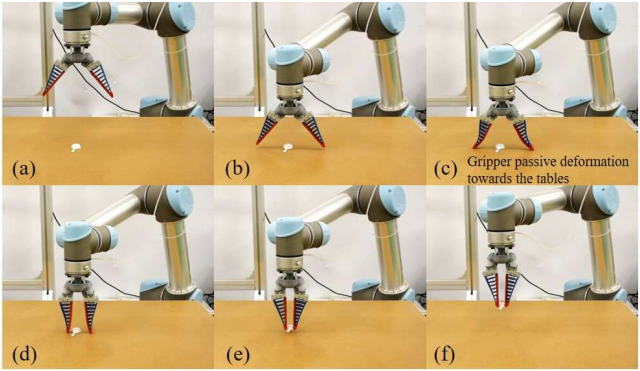

**Grasping primitive 1: Grasping flat/small objects**

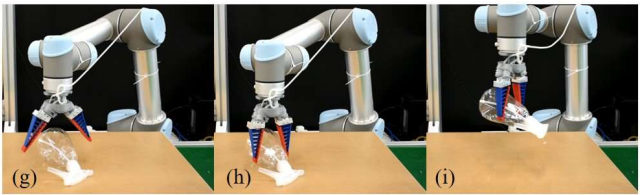

**Grasping primitive 2: Grasping normal sized objects**

Figure 4. Demonstration of the two grasping primitives. (a-f) shows the grasping primitive 1. (g-i) shows the grasping primitive 2.

compared our methods with state-of-the-art grasp detection methods. Thirdly, to highlight one of our contributions, we examined the height prediction output of our neural network.

To demonstrate our method's robustness over depth sensing, we wish to perform grasping on objects hardly visible or wrongly perceived by commercial RGBD cameras. [30] is commonly used in robotics experiments, but most items in [30] are opaque and visible in depth sensors. Because of lacking reproducible benchmark objects that challenge grasping due to perception uncertainty, we collected our own test objects from frequently used daily objects.

The set contains transparent objects, specular objects, flat objects of varying shapes and sizes. Some of them have multiple attributes (e.g., both flat and specular). All of the objects are different in geometry or appearance from the training set. In total, 24 household items are tested and 17 of which are depicted in Fig 3, where Fig 3(a) shows the RGB output of the camera and Fig 3(b) shows the depth output of the camera. It could tell that the depth sensing of these objects is quite uncertain and unclear.

### A. Physical components

The grasping system setup has been introduced in Section II(A) and depicted in Fig 1(a). Training and inference computations are performed on a PC running Ubuntu 18.04 with an NVIDIA GeForce RTX 2080 Ti and an Intel Core i7-10700. The code is predominantly written in Python. Motion planning was implemented using MoveIt. [38] It takes 10 mins to train the network on this platform.

### B. Fuzzy-depth Objects Grasping

To evaluate our grasping pipeline and network under the most common conditions, we performed real grasping on unseen fuzzy-depth objects. Five trials were performed on each object. For each trial, one object was randomly selected and thrown into the workspace shown in Fig. 4(a).

A successful trial is defined as if the object can be picked up and placed in a basket next to the workspace without slipping, dropping, or damaging of the objects. The overall grasp successful rate was 84.2% (101/120). Noting that our method achieved the best performance on flat objects without specularity or transparency (90%) and failed more on grasping objects with strong specularity (83%). Some typical failures are smooth objects slip during closing the fingers, predicting wrong in-plane rotation of grasping for extremely small objects.

### C. Performance Comparison

We compared our method and two other approaches using our collected object set and experimental setup. The first approach was GG-CNN2 with RGBD input and the second was the original depth-based GG-CNN2 [24]. All networks were trained using the same training dataset with the same hyperparameters. The results are shown in Table I. GG-CNN2 with RGBD input and our method significantly outperforms depth-based GG-CNN2. When depth sensing was uncertain, depth-based GG-CNN2 failed to distinguish the flat objects and got distracted by notable depth errors of specular and transparent objects. By adding color channels as input, RGBD-based GG-CNN2 started to detect more reliable grasps but failed to transfer the detected grasps into robot coordinate when the measured depth at a grasp point was missing or incorrect. In terms of speed, all networks in comparison were relatively small, which allowing them to generate grasping poses in real-time.

### D. Height Prediction

Although our method did not require accurate height estimation of objects, we found the height prediction module of our network works for most test objects. We performed 5 trials on each object in the test set that is unseen during training and examined the height prediction values of the best grasp. Ground truth was measured manually. The results are shown in Table II. The errors of most predictions were less than 20 mm. We noted that though at times our network has larger errors on some objects, it did not necessarily cause grasping failures thanks to the robustness of the soft grasping primitives.

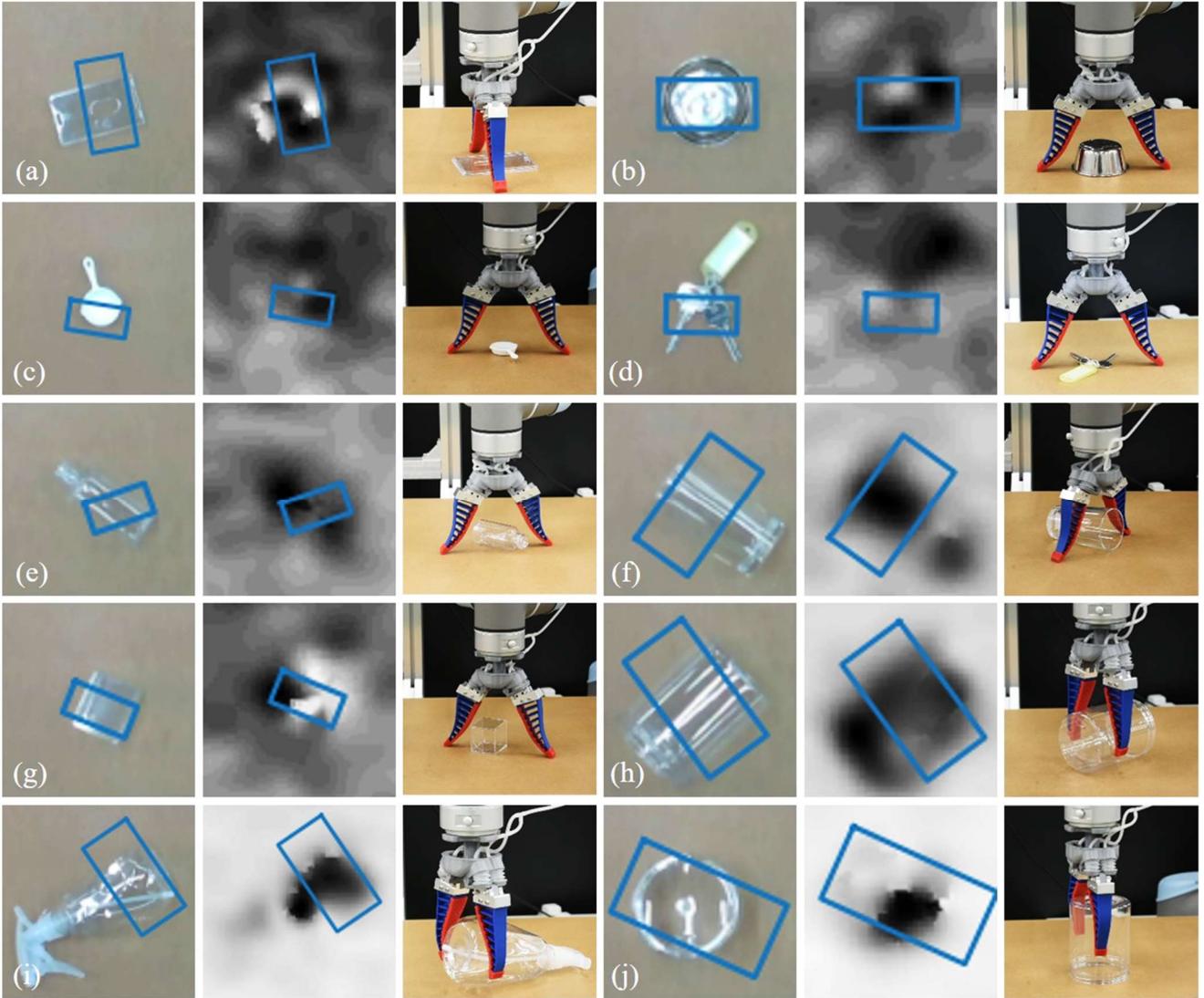

Figure 5. Qualitative results of grasping some Fuzzy-depth objects. The color and depth images from the overhead RGBD camera and a third picture showing the corresponding physical grasp (a). Flat and transparent card cover (b) Specular stainless bowl (c) Small pan-like measuring tool (d) Small and specular keys (e) Transparent bottle (f) Transparent glass (g) Transparent and specular pen holder (h)-(j) Transparent and specular objects with non-negligible height. In these cases, the gripper automatically adjusts grasping height based on the height of objects.

**Table I. Success Rates of Isolated Grasping**

| Method | Attempts | Failures | Success Rates (%) | Time (s) |
|---|---|---|---|---|
| FSG | 120 | 19 | 84 | 0.016 |
| RGBD-based GG-CNN2 | 120 | 28 | 76 | 0.015 |
| Depth-based GG-CNN2 | 120 | 68 | 43 | 0.016 |

**Table II. Height predictions averaged over 5 trials (mm)**

| Objects | Absolute error | $\sigma$ |
|---|---|---|
| Sprinkling can | 0.5 | 1.7 |
| Keys | 4.1 | 2.5 |
| Jigger | 20.0 | 0.6 |
| Pen holder | 2.8 | 2.2 |
| Mini bottle | 6.8 | 1.2 |
| Card cover | 41.0 | 1.8 |
| Jar 1 | 10.7 | 1.6 |
| Jar 2 | 5.1 | 1.1 |
| Spray bottle | 19.9 | 2.8 |
| Metal sealing cap | 17.6 | 0.9 |
| Stainless spoon | 1.1 | 3.1 |

## V. Conclusion and Future Work

In this paper, we present a grasping solution towards fuzzy-depth objects. The grasping system is majorly constructed by the FSG algorithm and a soft robotic hand. The algorithm leverages both color and depth images from a commercially available RGB-D camera and a neural network to predict grasps in 2D so that prediction is not confused by highly uncertain depth sensing of fuzzy-depth objects. Our neural network for grasp synthesis is small and therefore fast in computation. Two soft grasping primitives take the camera measured depth and neural network predicted depth into consideration jointly and determine the final grasps to be executed by our soft gripper in 2.5D. We noticed that our proposed FSG algorithm can exploit the merits of soft grippers, such as grasping compliance and adaptability.

In future work, we will make use of synthetic datasets or simulators so that labeling height will require less labor. Furthermore, grasping cluttered fuzzy-depth objects is also a valuable challenge to be solved.